\documentclass[a4paper,twoside]{article}

\usepackage{epsfig}
\usepackage{subcaption}
\usepackage{calc}
\usepackage{amssymb}
\usepackage{amstext}
\usepackage{amsmath}
\usepackage{amsthm}
\usepackage{multicol}
\usepackage{pslatex}
\usepackage{apalike}

\usepackage[outdir=./]{epstopdf}
\usepackage{algorithm}
\usepackage[noend]{algpseudocode}
\usepackage{nth}
\usepackage[all]{nowidow}
\usepackage{bbm}
\usepackage{booktabs,etoolbox}

\makeatletter
\def\algbackskip{\hskip-\ALG@thistlm}
\makeatother

\usepackage{array}
\newcolumntype{P}[1]{>{\centering\arraybackslash}p{#1}}

\algnewcommand{\algorithmicsubalgorithm}{\textbf{sub-algorithm}}
\algdef{SE}[SUBALG]{SubAlgorithm}{EndSubAlgorithm}{}{\algorithmicend\ \algorithmicsubalgorithm}%
\algtext*{EndSubAlgorithm}

\newcommand{\Image}{\textbf{I}}

\newcommand{\Objects}{{\mathcal{O}}}
\newcommand{\Object}[1]{\textbf{O}_{#1}}
\newcommand{\NumObjects}{N}
\newcommand{\MaxObjects}{N_{\max}}

\newcommand{\Prototypes}{\mathcal{P}}
\newcommand{\Prototype}[1]{\textbf{P}_{#1}}
\newcommand{\Masks}{\mathcal{M}}
\newcommand{\Mask}[1]{\textbf{M}_{#1}}
\newcommand{\NumPrototypes}{P}

\newcommand{\Templates}{\mathcal{T}}
\newcommand{\Template}[1]{\textbf{T}_{#1}}
\newcommand{\NumTemplates}{T}

\newcommand{\Locs}{\mathbf{L}}
\newcommand{\Disp}{\boldsymbol{\delta}_{x,y}}
\newcommand{\FFT}{\mathcal{F}}
\newcommand{\IFFT}{\mathcal{F}^{-1}}

\newcommand{\Error}{\textbf{E}}

\newcommand{\ComposeFunction}{\mathcal{G}}
\newcommand{\Loss}{\mathcal{L}}

\usepackage{subcaption}
\usepackage{hyperref}
\usepackage{caption}
\usepackage{booktabs}                        
\usepackage{tabularx}
\usepackage{multicol}

\def\R{{\mathbb R}}

\newcommand{\Figure}[1]{Figure~\ref{#1}}

\newcommand{\Equation}[1]{Equation~\eqref{#1}}

\newcommand{\Table}[1]{Table~\ref{#1}}

\newcommand{\Section}[1]{Section~\ref{#1}}

\newcommand{\Appendix}[1]{Appendix~\ref{#1}}
\newcommand{\Algorithm}[1]{Algorithm~\ref{#1}}

\usepackage{SCITEPRESS}     

\begin{document}

\title{Unsupervised Image Decomposition with Phase-Correlation Networks}

\author{\authorname{Angel Villar-Corrales\orcidAuthor{0000-0001-5805-2098} and Sven Behnke\orcidAuthor{0000-0002-5040-7525}}
\affiliation{Autonomous Intelligent Systems, University of Bonn, Germany}
\email{villar@ais.uni-bonn.de}
}


\abstract{
	The ability to decompose scenes into their object components is a desired property for autonomous agents, allowing them to reason and act in their surroundings. Recently, different methods have been proposed to learn object-centric representations from data in an unsupervised manner.
	These methods often rely on latent representations learned by deep neural networks, hence requiring high computational costs and large amounts of curated data. Such models are also difficult to interpret. 
	To address these challenges, we propose the \emph{Phase-Correlation Decomposition Network} (PCDNet), a novel model that decomposes a scene into its object components, which are represented as transformed versions of a set of learned object prototypes. The core building block in PCDNet is the \emph{Phase-Correlation Cell} (PC Cell), which exploits the frequency-domain representation of the images in order to estimate the transformation between an object prototype and its transformed version in the image.
	In our experiments, we show how PCDNet outperforms state-of-the-art methods for unsupervised object discovery and segmentation on simple benchmark datasets and on more challenging data, while using a small number of learnable parameters and being fully interpretable.
	Code and models to reproduce our experiments can be found in \url{https://github.com/AIS-Bonn/Unsupervised-Decomposition-PCDNet}.
}

\onecolumn \maketitle \normalsize \setcounter{footnote}{0} \vfill

\section{Introduction}

Humans understand the world by decomposing scenes into objects that can interact
with each other. Analogously, autonomous systems’ reasoning and scene understanding capabilities could benefit from decomposing scenes into objects and modeling each of these independently.
This approach has been proven beneficial to perform a wide variety of computer vision tasks without explicit supervision, including unsupervised object detection~\cite{Eslami_AttendInferRepeatSceneUnderstanding_2016}, future frame prediction~\cite{Weis_UnmaskingInductiveBiasesOfUnsupervisedObjectRepresentationsForVideoSequences_2020,Greff_IodineMultiObjectRepresentationLearningWithIterativeVariationalInference_2019}, and object tracking~\cite{He_TrackingByAnnimation_2019,Veerapaneni_EntityAbstractioninVisualModelBasedReinforcementLearning_2020}.

Recent works propose extracting object-centric representations without the need for explicit supervision through the use of deep variational auto-encoders~\cite{Kingma_AutoEncodingVariationalBayes_2013} (VAEs) with spatial attention mechanisms~\cite{Burgess_MonetUnsupervisedSceneDecompositionRepresentation_2019,crawford2019spatially}. 
However, training these models often presents several difficulties, such as long training times, requiring a large number of trainable parameters, or the need for large curated datasets. Furthermore, these methods suffer from the inherent lack of interpretability which is characteristic of deep neural networks (DNNs).

\begin{figure*}[t]
	\centering
	\includegraphics[width=0.92\linewidth]{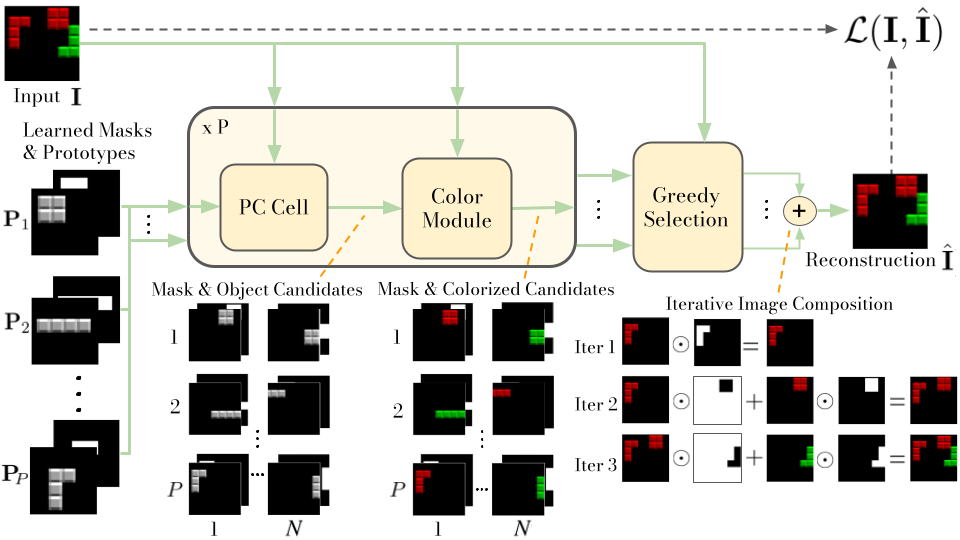}
	\caption{PCDNet decomposition framework. First, the Phase Correlation (PC) Cell estimates the $\NumObjects$ translation parameters that best align each learned prototype to the objects in the image, and uses them to obtain $(\NumPrototypes \times \NumObjects)$ object and mask candidates. Second, the color module assigns a color to each of the transformed prototypes. Finally, a greedy selection algorithm reconstructs the input image by iteratively combining the colorized object candidates that minimize the reconstruction error.}
	\label{fig:phacord}
\end{figure*}

To address the aforementioned issues, we propose a novel image decomposition framework: the \emph{\textbf{P}hase-\textbf{C}orrelation \textbf{D}ecomposition \textbf{Net}work } (PCDNet). Our method assumes that an image is formed as an arrangement of multiple objects, each belonging to one of a finite number of different classes.
Following this assumption, PCDNet decomposes an image into its object components, which are represented as transformed versions of a set of learned object prototypes.

The core building block of the PCDNet framework is the \emph{Phase Correlation Cell} (PC Cell). This is a differentiable module that exploits the frequency-domain representations of an image and a prototype to estimate the transformation parameters that best align a prototype to its corresponding object in the image.
The PC Cell localizes the object prototype in the image by applying the phase-correlation method~\cite{Alba_PahseCorrelationImageAlignment_2012}, i.e., finding the peaks in the cross-correlation matrix between the input image and the prototype. Then, the PC Cell aligns the prototype to its corresponding object in the image by performing the estimated phase shift in the frequency domain.

PCDNet is trained by first decomposing an image into its object components, and then reconstructing the input by recombining the estimated object components following the ``dead leaves'' model approach, i.e., as a superposition of different objects. 
The strong inductive biases introduced by the network structure allow our method to learn fully interpretable prototypical object-centric representations without any external supervision while keeping the number of learnable parameters small.
Furthermore, our method also disentangles the position and color of each object in a human-interpretable manner.

In summary, the contributions of our work are as follows:

\begin{itemize}
	\item We propose the PCDNet model, which decomposes an image into its object components, which are represented as transformed versions of a set of learned object prototypes. 
	
	\item Our proposed model exploits the frequency-domain representation of images so as to disentangle object appearance, position, and color without the need for any external supervision. 
	
	\item Our experimental results show that our proposed framework outperforms recent methods for joined unsupervised object discovery, image decomposition, and segmentation on benchmark datasets, while significantly reducing the number of learnable parameters, allowing for high throughput, and maintaining interpretability.
	
\end{itemize}

\section{Related Work}

\subsection{Object-Centric Representation Learning}

The field of representation learning~\cite{Bengio_RepresentationLearningReview_2013} has seen much attention in the last decade, giving rise to great advances in learning hierarchical representations~\cite{Paschalidou_LearningUnsupervisedPartDecompositionOf3DObjects_2020,Stanic_HierarchicalRelationalInference_2020} or in disentangling the underlying factors of variation in the data~\cite{Locatello_ChallengingAssumptionsInLearningOfDissentangledRepresentations_2019,Burgess_UnderstandingDisentanglingInBetaVAE_2018}.
Despite these successes, these methods often rely on learning representations at a scene level, rather than learning in an object-centric manner, i.e., simultaneously learning representations that address multiple, possibly repeating, objects.

In the last few years, several methods have been proposed to perform object-centric image decomposition in an unsupervised manner. 

A first approach to object-centric decomposition combines VAEs with attention mechanisms to decompose a scene into object-centric representations. The object representations are then decoded to reconstruct the input image.
These methods can be further divided into two different groups depending on the class of latent representations used. 
On the one hand, some methods~\cite{Eslami_AttendInferRepeatSceneUnderstanding_2016,Kosiorek_SequentialAttendInferRepeat_2018,Stanic_HierarchicalRelationalInference_2020,He_TrackingByAnnimation_2019} explicitly encode the input into factored latent variables, which represent specific properties such as appearance, position, and presence.
On the other hand, other models~\cite{Burgess_MonetUnsupervisedSceneDecompositionRepresentation_2019,Weis_UnmaskingInductiveBiasesOfUnsupervisedObjectRepresentationsForVideoSequences_2020,Locatello_ObjectCentricLearningWithSlotAttention_2020} decompose the image into unconstrained per-object latent representations.

Recently, several proposed methods~\cite{Greff_IodineMultiObjectRepresentationLearningWithIterativeVariationalInference_2019,Engelcke_GenesisGeneraticeSceneInferenceWithObjectCentricRepresentations_2019,Engelcke_GenesisV2InferringObjectRepresentationsWithoutIterativeRefinement_2021,Veerapaneni_EntityAbstractioninVisualModelBasedReinforcementLearning_2020,Lin_UnsupervisedObjectOrientedSceneRepresentationViaSpatialAttentionAndDecomposition_2020} use parameterized spatial mixture models with variational inference to decode object-centric latent variables.


Despite these recent advances in unsupervised object-centric learning, most existing methods rely on DNNs and attention mechanisms to encode the input images into latent representations, hence requiring a large number of learnable parameters and high computational costs. Furthermore, these approaches suffer from the inherent lack of interpretability characteristic of DNNs.
Our proposed method exploits the strong inductive biases introduced by our scene composition model in order to decompose an image into object-centric components without the need for deep encoders, using only a small number of learnable parameters, and being fully interpretable.

\subsection{Layered Models}
The idea of representing an image as a superposition of different layers
has been studied since the introduction of the ``dead leaves'' model by~\cite{matheron1968schema}.
This model has been extended to handle natural images and scale-invariant representations~\cite{Lee_OcclusionModelsForNaturalImages_2001}, as well as video sequences~\cite{Jojic_LearningFlexibleSpritesInVideoLayers_2001}.
%
More recently, several works~\cite{Yang_LRGan_2017,lin2018st,Zhang_CopyPasteGAN_2020,aksoy2017unmixing,Arandjelovic_ObjectDiscoveryWithCopyPasteGAN_2019,Sbai_UnuspervisedImageDecompositioninVectorLayers_2020} combine deep neural networks and ideas from layered image formation for different generative tasks, such as editing or image composition.
However, the aforementioned approaches are limited to foreground/background layered decomposition, or to represent the images with a small number of layers.

The work most similar to ours was recently presented  by~\cite{Monnier_UnsupervisedLayeredImageDecompositionIntoObjectPrototypes_2021}. 
The authors propose a model to decompose an image into overlapping layers, each containing an object from a predefined set of categories. The object layers are obtained with a cascade of spatial transformer networks, which learn transformations that align certain object sprites to the input image.

While we also follow a layered image formation, our PCDNet model is not limited to a small number of layers, hence being able to represent scenes with multiple objects. PCDNet represents each object in its own layer, and uses learned alpha masks to model occlusions and superposition between layers.

\subsection{Frequency-Domain Neural Networks}

Signal analysis and manipulation in the frequency domain is one of the most widely used tools in the field of signal processing~\cite{Proakis_DigitalSignalProcessing_2004}. However, frequency-domain methods are not so developed for solving computer vision tasks with neural networks. They mostly focus on specific applications such as compression~\cite{Xu_LearningInTheFrequencyDomain_2020,Gueguen_FasterNeuralNetworkStraightFromJPEG_2018}, image super-resolution and denoising~\cite{Fritsche_FrequencySeparationForRealWorldSuperResolution_2019,Villar_DeepLearningDesignsForSuperResolutionNoisyImages_2021,Kumar_ConvNeuralNetworksForWaveletDomainSuperResolution_2017}, or accelerating the calculation of convolutions~\cite{Mathieu_FastTrainingOfCNNsThroughFFTs_2013,Ko_EnergyEfficientTrainingCNNFrequencyDomain_2017}.

In recent years, a particular family of frequency-domain neural networks---the \emph{phase-correlation networks}---has received interest from the research community and has shown promise for tasks such as future frame prediction~\cite{Farazi_LocalFrequencyDomainTransformerNwtworksForVideoPrediction_2021,Wolter_ObjectCenteredFourierMotionEstimation_2020} and motion segmentation~\cite{Farazi_MotionSegmentationUsingFrequencyDomainTransformerNetworks_2020}. 
Phase-correlation networks compute normalized cross-correlations in the frequency domain and operate with the phase of complex signals in order to estimate motion and transformation parameters between consecutive frames, which can be used to obtain accurate future frame predictions requiring few learnable parameters. 
Despite these recent successes, phase-correlation networks remain unexplored beyond the tasks of video prediction and motion estimation. Our proposed method presents a first attempt at applying phase correlation networks for the tasks of scene decomposition and unsupervised object-centric representation learning.

\begin{figure*}[t]	
	\begin{subfigure}{0.55\textwidth}
		\centering
		\includegraphics[width=1.0\linewidth]{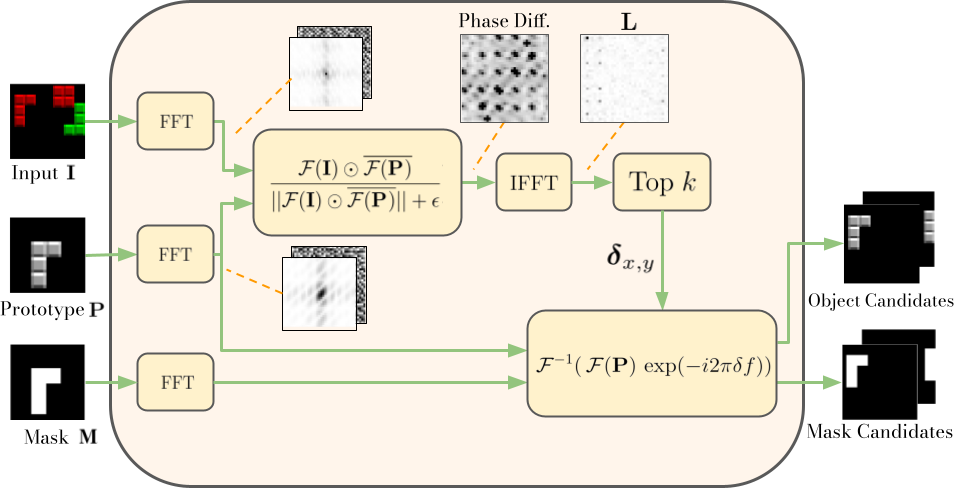}
		\caption{PC Cell}
		\label{fig: pccell}
	\end{subfigure}
	\hspace{0.0\textwidth}
	\begin{subfigure}{0.45\textwidth}
		\centering
		\vspace{0.65cm}
		\includegraphics[width=1.0\linewidth]{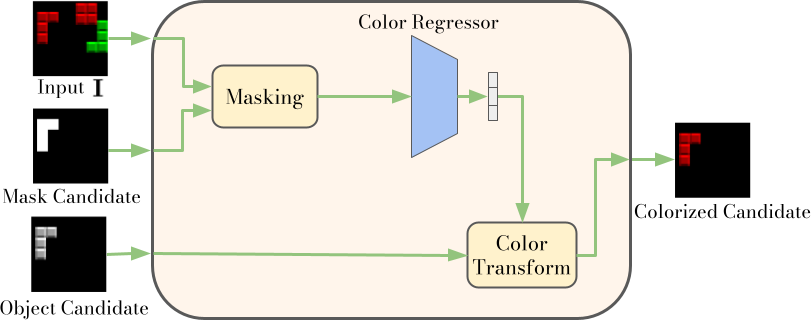}
		\vspace{0.2cm}
		\caption{Color Module}
		\label{fig:color module}
	\end{subfigure}
	\caption{\textbf{(a)}: Inner structure of the PC Cell. First, the translation parameters are estimated by finding the correlation peaks between the object prototype and the input image. Second, the prototype is shifted by phase shifting in the frequency domain. \textbf{(b)}: The \emph{Color Module} estimates color parameters from the input and aligns the color channels of a translated object prototype.}
	\label{a}
\end{figure*}

\section{Phase-Correlation Decomposition Network}

In this section, we present our image decomposition model: PCDNet. Given an input image $\Image$, PCDNet aims at its decomposition into $\NumObjects$ independent objects $\Objects = \{\Object{1}, \Object{2}, ..., \Object{\NumObjects}\}$.
In this work, we assume that these objects belong to one out of a finite number $\NumPrototypes$ of classes, and that there is a known upper bound to the total number of objects present in an image ($\MaxObjects$).

Inspired by recent works in prototypical learning and clustering~\cite{Li_PrototypicalContrastiveLearningOfUnsupervisedRepresentations_2020,Monnier_DeepTransformationInvariantClustering_2020}, we design our model such that the objects in the image can be represented as transformed versions of a finite set of object prototypes $\Prototypes = \{\Prototype{1}, \Prototype{2}, ..., \Prototype{\NumPrototypes}\}$. 
Each object prototype $\Prototype{i} \in \R^{H,W}$ is learned along with a corresponding alpha mask $\Mask{i} \in \R^{H,W}$, which is used to model occlusions and superposition of objects.
Throughout this work, we consider object prototypes to be in gray-scale and of smaller size than the input image.
PCDNet simultaneously learns suitable object prototypes, alpha masks and transformation parameters in order to accurately decompose an image into object-centric components.

An overview of the PCDNet framework is displayed in \Figure{fig:phacord}. First, the \emph{PC Cell} (\Section{subsec: pc-cell}) estimates the candidate transformation parameters that best align the object prototypes to the objects in the image, and generates object candidates based on the estimated parameters. Second, a \emph{Color Module} (\Section{subsec: color module}) transforms the object candidates by applying a learned color transformation. Finally, a \emph{greedy selection algorithm} (\Section{subsec: greedy selection}) reconstructs the input image by iteratively selecting the object candidates that minimize the reconstruction error.
%

\subsection{Phase-Correlation Cell}
\label{subsec: pc-cell}

The first module of our image decomposition framework is the PC Cell, as depicted in \Figure{fig:phacord}. This module first estimates the regions of an image where a particular object might be located, and then shifts the prototype to the estimated object location.
Inspired by traditional image registration methods~\cite{Reddy_FFTBasedImageRegistration_1996,Alba_PahseCorrelationImageAlignment_2012}, we adopt an approach based on phase correlation. 
This method estimates the relative displacement between two images by computing the normalized cross-correlation in the frequency domain.

Given an image $\Image$ and an object prototype $\Prototype{}$, the PC Cell first transforms both inputs into the frequency domain using the \emph{Fast Fourier Transform} (FFT, $\FFT$).
Second, it computes the phase differences between the frequency representations of image and prototype, which can be efficiently computed as an element-wise division in the frequency domain. Then, a localization matrix $\Locs$ is found by applying the inverse FFT ($\IFFT$) on the normalized phase differences: 

\vspace{-0.2cm}
\begin{align}
	& \Locs = \IFFT \Big( \frac{\FFT(\Image) \odot \overline{\FFT(\Prototype{})}}  {|| \FFT(\Image) \odot \overline{\FFT(\Prototype{})} || + \epsilon} \Big), 
	\label{eq:phase corr}
\end{align}

where $\overline{\FFT(\Prototype{})}$ denotes the complex conjugate of $\FFT(\Prototype{})$, $\odot$ is the Hadamard product, $||\cdot||$ is the modulus operator, and $\epsilon$ is a small constant to avoid division by zero.
Finally, the estimated relative pixel displacement ($\Disp = (\delta_x, \delta_y)$) can then be found by locating the correlation peak in $\Locs$:

\vspace{-0.4cm}
\begin{align}
	& \Disp = \arg \max (\Locs) \; .
\end{align}

In practical scenarios, we do not know in advance which objects are present in the image or whether there are more than one objects from the same class. To account for this uncertainty, we pick the largest $\MaxObjects$ correlation values from $\Locs$ and consider them as candidate locations for an object.

Finally, given the estimated translation parameters, the PC Cell relies on the Fourier shift theorem to align the object prototypes and the corresponding alpha masks to the objects in the image. Given the translation parameters $\delta_x$ and $\delta_y$, an object prototype is shifted using

\begin{align}
	& \Template{} = \IFFT( \FFT(\Prototype{}) \exp(-i 2 \pi (\delta_x \mathbf{f}_x + \delta_y \mathbf{f}_y )), 
\end{align}

where $\mathbf{f}_x$ and $\mathbf{f}_y$ denote the frequencies along the horizontal and vertical directions, respectively. 

\Figure{fig: pccell} depicts the inner structure of the PC Cell, illustrating each of the phase correlation steps and displaying some intermediate representations, including the magnitude and phase components of each input, the normalized cross-correlation matrix, and the localization matrix $\Locs$.

\subsection{Color Module}
\label{subsec: color module}

The PC Cell module outputs translated versions of the object prototypes and their corresponding alpha masks. However, these translated templates need not match the color of the object represented in the image. This issue is solved by the \emph{Color Module}, which is illustrated in \Figure{fig:color module}. It learns color parameters from the input image, and transforms the translated prototypes according to the estimated color parameters.

Given the input image and the translated object prototype and mask, the color module first obtains a masked version of the image containing only the relevant object. This is achieved through an element-wise product of the image with the translated alpha mask.
The masked object is fed to a neural network, which learns the color parameters (one for gray-scale and three for RGB images). Finally, these learned parameters are applied to the translated object prototypes with a channel-wise affine transform.
Further details about the color module are given in \Appendix{app: color module}.

\subsection{Greedy Selection}
\label{subsec: greedy selection}

The PC Cell and color modules produce  $\NumTemplates = \MaxObjects \times \NumPrototypes$ translated and colorized candidate objects ($\Templates=\{\Template{1}, ..., \Template{\NumTemplates}\}$) and their corresponding translated alpha masks ($\{\Mask{1}, ..., \Mask{\NumTemplates}\}$).
The final module of the PCDNet framework selects, among all candidates, the objects that minimize the reconstruction error with respect to the input image. 

The number of possible object combinations grows exponentially with the maximum number of objects and the number of object candidates ($\NumTemplates^{\MaxObjects}$), which quickly makes it infeasible to evaluate all possible combinations.
Therefore, similarly to \cite{Monnier_UnsupervisedLayeredImageDecompositionIntoObjectPrototypes_2021}, we propose a greedy algorithm that selects in a sequential manner the objects that minimize the reconstruction loss. 
The greedy nature of the algorithm reduces the number of possible object combinations to $\NumTemplates \times \MaxObjects$, hence scaling to images with a large number of objects and prototypes.

The greedy object selection algorithm operates as follows.
At the first iteration, we select the object that minimizes the reconstruction loss with respect to the input, and add it to the list of selected objects. Then, for each subsequent iteration, we greedily select the object that, combined with the previously selected ones, minimizes the reconstruction error.
This error is computed using \Equation{eq: error}, which corresponds to the mean squared error between the input image ($\Image$) and a combination of the selected candidates ($\ComposeFunction(\Templates)$).

The objects are combined recursively in an overlapping manner, as shown in \Equation{eq: proto combination}, so that the first selected object ($\Template{1}$) corresponds to the one closest to the viewer, whereas the last selected object ($\Template{\NumObjects}$) is located the furthest from the viewer:
\begin{align}
	\Error(\Image, \Templates)  = \; & ||\Image - \ComposeFunction(\Templates)||_2^2 
	\label{eq: error}  \\
	\ComposeFunction(\Templates)  = \; & \Template{i+1} \odot (1 - \Mask{i}) + \Template{i} \odot \Mask{i} \nonumber \\
	&\forall \, i \, \in \, \{\NumObjects-1, ..., 1\}   \label{eq: proto combination} . 
\end{align}

An example of this image composition is displayed in \Figure{fig:phacord}. 
This reconstruction approach inherently models relative depths, allowing for a simple, yet effective, modeling of occlusions between objects.

\subsection{Training and Implementation Details}
\label{subsec: training}

%
%

We train PCDNet in an end-to-end manner to reconstruct an image as a combination of transformed object prototypes. The training is performed by minimizing the reconstruction error ($\Loss_{MSE}$), while regularizing the prototypes to with respect to the $\ell_1$ norm ($\Loss_{L1}$), and enforcing smooth alpha masks with a total variation regularizer~\cite{Rudin_TotalVariationBasedImageRestoration_1994} ($\Loss_{TV}$).
Specifically, we minimize the following loss function:

\vspace{-0.3cm}

\begin{align}
    \Loss_{} &= \Loss_{MSE} +  \lambda_{L1} \; \Loss_{L1} +  \lambda_{TV} \; \Loss_{TV} 
	\label{eq: loss function} \\
	\Loss_{MSE} &= ||\Image - \ComposeFunction(\Templates')||_2^2 \\
	\Loss_{L1} &= \frac{1}{\NumPrototypes} \sum_{\Prototype{} \in \Prototypes}  ||\Prototype{}||_1 \\
	\Loss_{TV} &= \frac{1}{\NumPrototypes} \sum_{\Mask{} \in \Masks} \sum_{i,j} |\Mask{i+1,j} - \Mask{i,j}| + |\Mask{i,j+1} - \Mask{i,j}|
\end{align}

where $\Templates'$ corresponds to the object candidates selected by the greedy selection algorithm,
$\Prototypes$ are the learned object prototypes, and $\Masks$ to the corresponding alpha masks.
Namely, minimizing \Equation{eq: loss function} decreases the reconstruction error between the combination of selected object candidates ($\ComposeFunction(\Templates')$) and the input image, while keeping the object prototypes compact, and the alpha masks smooth.

\begin{table*}[t!]
	\centering
	\captionof{table}{Object discovery evaluation results on the Tetrominoes dataset. PCDNet outperforms SOTA methods, while using a small number of learned parameters. Moreover, our PCDNet has the highest throughput out of all evaluated methods. For each metric, the best result is highlighted in boldface, whereas the second best is underlined.}
	\label{table: ari eval}
	\vspace{-0.0cm}
	\normalsize
	\begin{tabular}{ p{5.3cm} | P{2.cm}  | P{2.3cm}  P{2.3cm}}    
		\midrule[2pt]
		\textbf{Model} & \textbf{ARI (\%) $\uparrow$} & \textbf{ Params $\downarrow$}  & \textbf{Imgs/s $\uparrow$} \\
		\midrule[1.4pt]
		
		Slot MLP~\cite{Locatello_ObjectCentricLearningWithSlotAttention_2020} &  35.1  &  -- & -- \\
		
		Slot Attention~\cite{Locatello_ObjectCentricLearningWithSlotAttention_2020} &  99.5  &  \underline{229,188} &  18.36\\
		
		ULID~\cite{Monnier_UnsupervisedLayeredImageDecompositionIntoObjectPrototypes_2021}  &  \underline{99.6}  &  659,755  & \underline{52.31} \\
		
		IODINE~\cite{Greff_IodineMultiObjectRepresentationLearningWithIterativeVariationalInference_2019} &  99.2  &  408,036 & 16.64\\
		
		PCDNet (ours) &  \textbf{99.7} &  \textbf{28,130}  & \textbf{58.96}  \\
				
		\midrule[2pt]
	\end{tabular}
\end{table*}

\begin{figure}[t]
	\centering
	\includegraphics[width=0.99\linewidth]{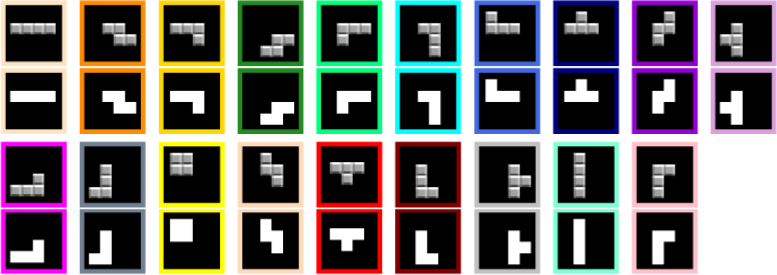}
	\caption{Object prototypes (top) and segmentation alpha masks (bottom) learned on the Tetrominoes dataset. Our model is able to discover in an unsupervised manner all 19 pieces.}
	\label{fig: tetris}
\end{figure}

In our experiments, we noticed that the initialization and update strategy of the object prototypes is of paramount importance for the correct performance of the PCDNet model. 
The prototypes are initialized with a small constant value (e.g., 0.2), whereas the center pixel is assigned an initial value of one, enforcing the prototypes to emerge centered in the frame.

During the first training iterations, we notice that the greedy algorithm selects some prototypes with a higher frequency that others, hence learning much faster.
In practice, this prevents other prototypes from learning relevant object representations, since they are not updated often enough.
To reduce the impact of uneven prototype discovery, we add, with a certain probability, some uniform random noise to the prototypes during the first training iterations. This prevents the greedy algorithm from always selecting, and hence updating, the same object prototypes and masks.

In datasets with a background, we add a special prototype to model a static background. 
In these cases, the input images are reconstructed by overlapping the objects selected by the greedy algorithm on top of the background prototype.
This background prototype is initialized by averaging all training images, and its values are refined during training.

\section{Experimental Results}

In this section, we quantitatively and qualitatively evaluate our PCDNet framework for the tasks of unsupervised object discovery and segmentation. 
PCDNet is implemented in Python using the PyTorch framework~\cite{Paszke_AutomaticDifferneciationInPytorch_2017}. 
A detailed report of the hyper-parameters used is given in \Appendix{subsec: Model and Hyper-Parameter Details}.

\begin{figure*}[t!]
	\hspace{0.02\textwidth}
	\begin{subfigure}{0.11\textwidth}
		\centering
		\includegraphics[width=1.0\linewidth]{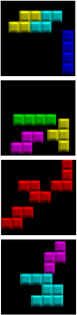}
		\caption{}
		\label{fig: orig}
	\end{subfigure}
	\hspace{0.01\textwidth}
	\begin{subfigure}{0.11\textwidth}
		\centering
		\includegraphics[width=1.0\linewidth]{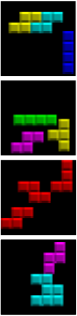}
		\caption{}
		\label{fig: recons}
	\end{subfigure}
	\hspace{0.01\textwidth}
	\begin{subfigure}{0.11\textwidth}
		\centering
		\includegraphics[width=1.01\linewidth]{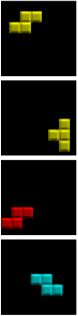}
		\caption{}
		\label{fig: obj1}
	\end{subfigure}
	\hspace{0.01\textwidth}
	\begin{subfigure}{0.11\textwidth}
		\centering
		\includegraphics[width=1.01\linewidth]{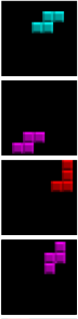}
		\caption{}
		\label{fig: obj2}
	\end{subfigure}
	\hspace{0.01\textwidth}
	\begin{subfigure}{0.11\textwidth}
		\centering
		\includegraphics[width=1.0\linewidth]{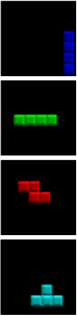}
		\caption{}
		\label{fig: obj3}
	\end{subfigure}
	\hspace{0.01\textwidth}
	\begin{subfigure}{0.11\textwidth}
		\centering
		\includegraphics[width=.99\linewidth]{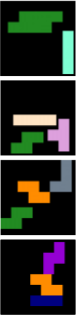}
		\caption{}
		\label{fig: segmentation}
	\end{subfigure}
	\hspace{0.01\textwidth}
	\begin{subfigure}{0.11\textwidth}
		\centering
		\includegraphics[width=.99\linewidth]{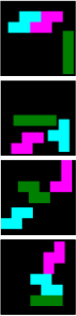}
		\caption{}
		\label{fig: instance}
	\end{subfigure}
	\hspace{0.02\textwidth}
	\vspace{-0.0cm}
	\caption{Qualitative decomposition and segmentation results on the Tetrominoes dataset. Last row shows a failure case. \textbf{(a)}: Original image. \textbf{(b)}: PCDNet Reconstruction. \textbf{(c)-(e)}: Colorized transformed object prototypes.  \textbf{(f)}: Semantic segmentation masks. Colors correspond to the prototype frames in \Figure{fig: tetris}. \textbf{(g)}: Instance segmentation masks.}
	\label{fig: qualitative tetris}
\end{figure*}

\subsection{Tetrominoes Dataset}
\label{subsec: Image Decomposition}

We evaluate PCDNet for image decomposition and object discovery on the Tetrominoes dataset~\cite{Greff_IodineMultiObjectRepresentationLearningWithIterativeVariationalInference_2019}. This dataset contains 60.000 training images and 320 test images of size $35 \times 35$, each composed of three non-overlapping Tetris-like sprites over a black background. The sprites belong to one out of 19 configurations and have one of six random colors.

\Figure{fig: tetris} displays the 19 learned object prototypes and their corresponding alpha masks from the Tetrominoes dataset. We clearly observe how PCDNet accurately discovers the shape of the different pieces and their tiled texture.

\Figure{fig: qualitative tetris} depicts qualitative results for unsupervised object detection and segmentation. 
In the first three rows, PCDNet successfully decomposes the images into their object components and precisely segments the objects into semantic and instance masks. The bottom row shows an example in which the greedy selection algorithm leads to a failure case.

For a fair quantitative comparison with previous works, we evaluate our PCDNet model for object segmentation using the Adjusted Rand Index~\cite{Hubert_ComparingPartitionsARI_1985} (ARI) on the  ground truth foreground pixels. ARI is a clustering metric that measures the similarity between two set assignments, ignoring label permutations, and ranges from 0 (random assignment) to 1 (perfect clustering).
We compare the performance of our approach with several existing methods: Slot MLP and Slot Attention~\cite{Locatello_ObjectCentricLearningWithSlotAttention_2020}, IODINE~\cite{Greff_IodineMultiObjectRepresentationLearningWithIterativeVariationalInference_2019} and Unsupervised Layered Image Decomposition~\cite{Monnier_UnsupervisedLayeredImageDecompositionIntoObjectPrototypes_2021} (ULID).

\Table{table: ari eval} summarizes the evaluation results for object discovery on the Tetrominoes dataset. From \Table{table: ari eval}, we see how PCDNet outperforms SOTA models, achieving 99.7\% ARI on the Tetrominoes dataset. PCDNet uses only a small percentage of learnable parameters compared
to other methods (e.g., only 6\% of the parameters from IODINE), and has the highest inference
throughput (images/s). Additionally, unlike other approaches, PCDNet obtains disentangled representations for the object appearance, position, and color in a human-interpretable manner.

\begin{figure}[t!]
	\centering
	\includegraphics[width=1.0\linewidth]{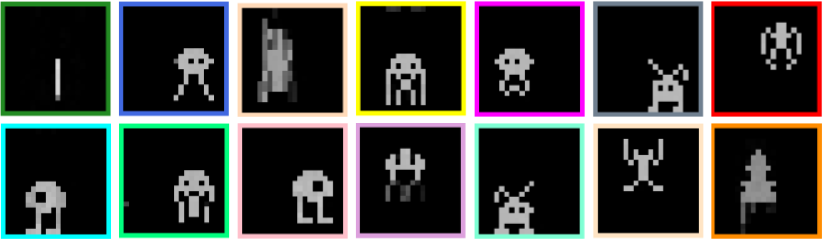}
	\caption{Object prototypes learned on the Space Invaders dataset. PCDNet discovers prototypes corresponding to the different elements from the game, e.g., aliens and ships.}
	\label{fig:prototypes atari}
\end{figure}

\subsection{Space Invaders Dataset}
\label{subsec: Space Invaders Dataset}

In this experiment, we use replays from humans playing the Atari game \emph{Space Invaders}, extracted from the Atari Grand Challenge dataset~\cite{Kurin_TheAtariGranChallengeDataset_2017}. 
PCDNet is trained to decompose the Space Invaders images into 50 objects, belonging to one of 14 learned prototypes of size $20 \times 20$. 

\begin{figure*}[t!]
	\centering
	\includegraphics[width=0.91\linewidth]{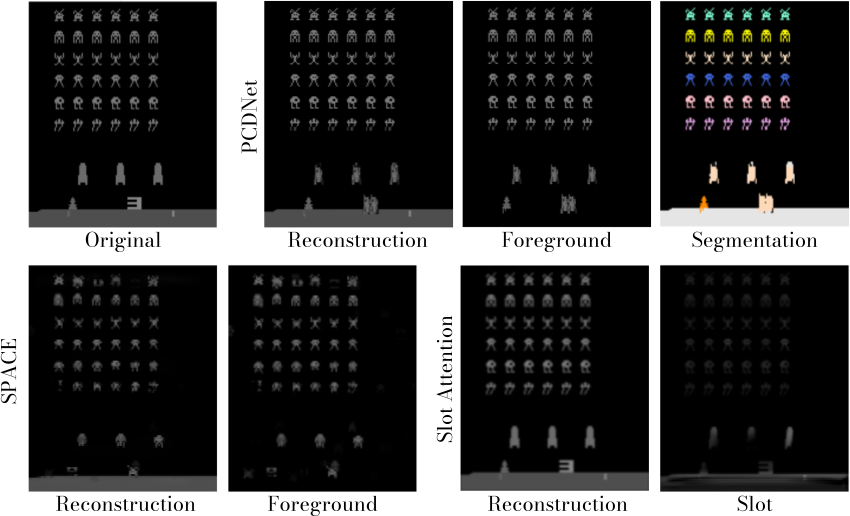}
	\caption{Comparison of different object-centric models on the Space Invaders dataset. PCDNet is the only one among the compared methods which successfully decomposes the image into accurate object components, and that has semantic knowledge of the objects. The color of each object corresponds to the frame of the corresponding prototype in \Figure{fig:prototypes atari}.}
	\label{fig:comp atari}
\end{figure*}

\Figure{fig:comp atari} depicts a qualitative comparison between our PCDNet model with SPACE~\cite{Lin_UnsupervisedObjectOrientedSceneRepresentationViaSpatialAttentionAndDecomposition_2020} and Slot Attention~\cite{Locatello_ObjectCentricLearningWithSlotAttention_2020}.

Slot Attention achieves an almost perfect reconstruction of the input image. However, it fails to decompose the image into its object components, uniformly scattering the object representations across different slots. In \Figure{fig:comp atari} (subplot labeled as \emph{Slot}) we show how one of the slots simultaneously encodes information from several different objects.
SPACE successfully decomposes the image into different object components, which are recognized as foreground objects. Nevertheless, the reconstructions appear blurred and several objects are not correct.
PCDNet achieves the best results among all compared methods. Our model successfully decomposes the input image into accurate object-centric representations. Additionally, PCDNet learns semantic understanding of the objects. 
\Figure{fig:comp atari} depicts a segmentation of an image from the Space Invaders dataset.
Further qualitative results on the Space Invaders dataset are reported in \Appendix{section: further qualitative results}.

\begin{figure*}[h!]
	\centering
	\includegraphics[width=0.91\linewidth]{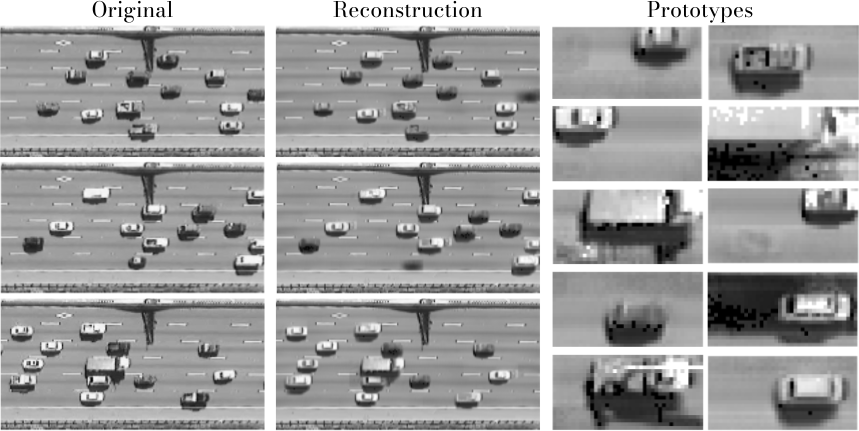}
	\caption{Object discovery on the NGSIM dataset. PCDNet learns different vehicle prototypes in an unsupervised manner.}
	\label{fig:cars}
\end{figure*}

\subsection{NGSIM Dataset}
\label{subsec: NGSIM Dataset}

In this third experiment, we apply our PCDNet model to discover vehicle prototypes from real traffic camera footage from the Next Generation Simulation (NGSIM) dataset~\cite{NGSIM_Dataset}.
We decompose each frame into up to 33 different objects, belonging to one of 30 learned vehicle prototypes.

\Figure{fig:cars} depicts qualitative results on the NGSIM dataset. 
We see how PCDNet is applicable to real-world data, accurately reconstructing the input image, while learning prototypes for different types of vehicles.
Interestingly, we notice how PCDNet learns the car shade as part of the prototype. This is a reasonable observation, since the shades are projected towards the bottom of the image throughout the whole video.

\section{Conclusion}

We proposed PCDNet, a novel image composition model that decomposes an image, in a fully unsupervised manner, into its object components, which are represented as transformed versions of a set of learned object prototypes. 
PCDNet exploits the frequency-domain representation of images to estimate the translation parameters that best align the prototypes to the objects in the image. 
The structured network used by PCDNet allows for an interpretable image decomposition, which disentangles object appearance, position and color without any external supervision. 
In our experiments, we show how our proposed model outperforms existing methods for unsupervised object discovery and segmentation on a benchmark synthetic dataset, while significantly reducing the number of learnable parameters, having a superior throughput, and being fully interpretable. Furthermore, we also show that the PCDNet model can also be applied for unsupervised prototypical object discovery on more challenging synthetic and real datasets.
%
%
We hope that our work paves the way towards further research on phase correlation networks for unsupervised object-centric representation learning.

\section*{ACKNOWLEDGMENTS}

This work was funded by grant BE 2556/18-2 (Research Unit FOR 2535
Anticipating Human Behavior) of the German Research Foundation (DFG).

%
%
%
\bibliographystyle{apalike}
{\small
	\bibliography{referencesAngel}
}

\begin{figure*}[t!]
	\centering
	\includegraphics[width=0.99\linewidth]{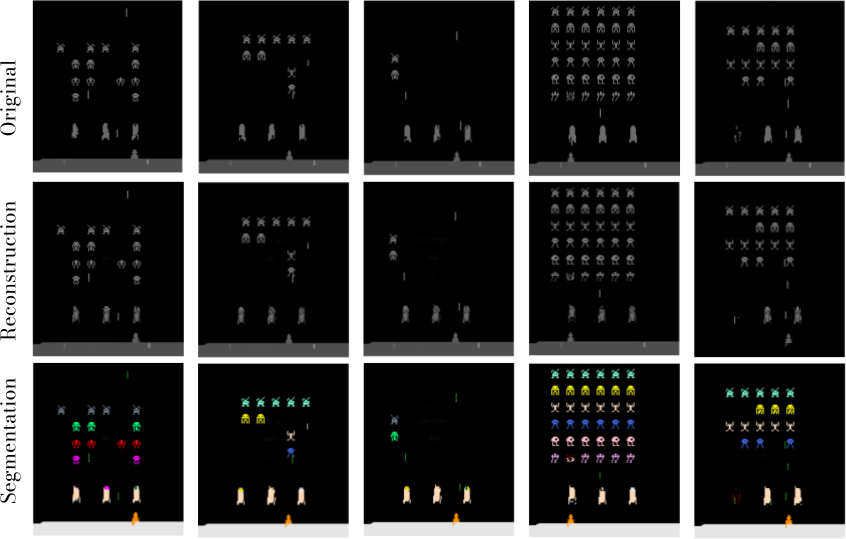}
	\caption{Additional PCDNet unsupervised segmentation qualitative results on the Space Invaders dataset.}
	\label{fig:atarisegment}
\end{figure*}

\appendix

\section{Model and Training Details}
\label{subsec: Model and Hyper-Parameter Details}

\begin{table}[t]
	\centering
	\caption{Hyper-parameter values used for each dataset.}
	\label{table: hyper-params}
	\vspace{-0.1cm}
	\normalsize
	
	\begin{tabular}{p{1.45cm} | P{1.4cm} P{1.4cm} P{1.2cm}}    
		\midrule[2pt]
		\textbf{Param.} & \textbf{Tetrominoes} & \textbf{Space Invaders} & \textbf{NGSIM} \\
		\midrule[1.4pt]
		LR & $0.003$ & $0.001$ & $0.013$\\
		Scheduler & 0.1 / 5 &  0.1 / 5 &  0.6 / 2 \\
		$\lambda_{L1}$ & $ 10^{-3}$ & 0 & $ 10^{-5}$ \\
		$\lambda_{TV}$ & $ 10^{-3}$ & 0 & $ 10^{-2}$ \\
		Batch size & 64 & 3 & 3 \\
		\midrule[2pt]
	\end{tabular}
\end{table}

\subsection{Training Details} We train all our experiments with an NVIDIA RTX 3090 GPU with 24 GB RAM using the Adam~\cite{Kingma_Adam_2014} update rule. Additionally, we use a learning rate scheduler that linearly decreases the learning rate by certain factor every few epochs.
We determine the values of our hyper-parameters using Optuna~\cite{Akiba_Optuna_2019}\footnote{Hyper-parameter ranges and further details can be found in \url{https://github.com/AIS-Bonn/Unsupervised-Decomposition-PCDNet}}.
The selected hyper-parameter values for each dataset are listed on \Table{table: hyper-params}. We report the learning rate (LR), learning rate scheduler parameters (LR factor / epochs), batch size, and regularizer weights ($\lambda_{L1}$ and $\lambda_{TV}$). 

The object prototypes are initialized with a constant value of 0.2 and with the center pixel set to one. This enforces the object prototypes to emerge centered. 
To prevent the greedy algorithm from always selecting the same prototypes during the first iterations, we add uniform random noise $\mathcal{U}[-0.5,0.5)$ to the prototypes with a probability of 80\%.

\begin{table}[t]
	\centering
	\caption{Implementation details of the \emph{Color Module} CNN.}
	\label{table: color module architecture}
	\vspace{-0.1cm}
	\normalsize
	
	\begin{tabular}{ p{3.5cm} P{2.0cm}  }    
		\midrule[2pt]
		\textbf{Layer} & \textbf{Dimension} \\
		\midrule[1.4pt]
		Input    &   $3 \times H \times W$   \\
		Conv. $(3 \times 3)$ + ReLU & $12 \times H \times W$ \\
		Batch Norm. & $12 \times H \times W$ \\
		Conv. $(3 \times 3)$ + ReLU & $12 \times H \times W$ \\
		Batch Norm. & $12 \times H \times W$ \\
		Global Avg. Pooling & $12 \times 1 \times 1$  \\
		Flatten & $12$  \\
		Fully Connected & 3 \\
		\midrule[2pt]
	\end{tabular}
\end{table}

\subsection{Color Module}
\label{app: color module}
The color module, depicted in \Figure{fig:color module}, is implemented in a similar fashion to a Spatial Transformer Network~\cite{Jaderberg_SpatialTransformerNetworks_2015} (STN). 
The masked image is fed to a neural network, which extracts certain color parameters corresponding to the masked object.
The architecture of this network is summarized in \Table{table: color module architecture}.
The extracted color parameters are applied to the translated object prototypes with a channel-wise affine transform.
Our color module shares similarities with other color transformation approaches~\cite{Kosiorek_StackedCapsuleAutoencoders_2019}. Despite applying the same affine channel transform, our method differs in the way the color parameters are computed.

\begin{algorithm}[b]
	\caption{Greedy Selection Algorithm}
	\label{algorithm: greedy selection algorithm}
	\begin{algorithmic}
		\Procedure{Greedy Selection Algorithm}{}
		\BState \emph{Inputs}:
		\State $\Image \gets \textit{Input image}$
		\State $\Templates = [\Template{1}, ..., \Template{\NumTemplates}] 
		\gets \textit{Object Candidates}$
		\State $\MaxObjects \gets \textit{Max. number of objects in the image}$
		\BState \emph{Returns}:
		\State $\Objects = [\Object{1}, ..., \Object{\MaxObjects}]  \gets \textit{Selected objects}$	
		
		\BState \emph{Algorithm}:
		\State $\Objects \gets [~]$
		
		\SubAlgorithm{Repeat $\MaxObjects$ times:}
		\State $\Error \gets [~]$
		\For{ $t$ in range [1, $\NumTemplates$]}
		
		\State $\Error_t = ||\Image - \ComposeFunction(\Objects, \Template{t})||_2^2 $
		
		\EndFor
		
		\State $q = \arg\min (\Error)$
		\State $\Objects \gets [\Objects, \Template{q}]$
		
		\EndSubAlgorithm
		
		\State return $\Objects$
		
		\EndProcedure
	\end{algorithmic}
\end{algorithm}

\subsection{Greedy Selection Algorithm}
\label{app: greedy algorithm}
\Algorithm{algorithm: greedy selection algorithm} illustrates the greedy selection algorithm used to select the colorized object candidates that best reconstruct the input image.

\section{Qualitative Results}
\label{section: further qualitative results}

\begin{figure*}[t!]	
	\begin{subfigure}{0.495\textwidth}
		\centering
		\includegraphics[width=1.0\linewidth]{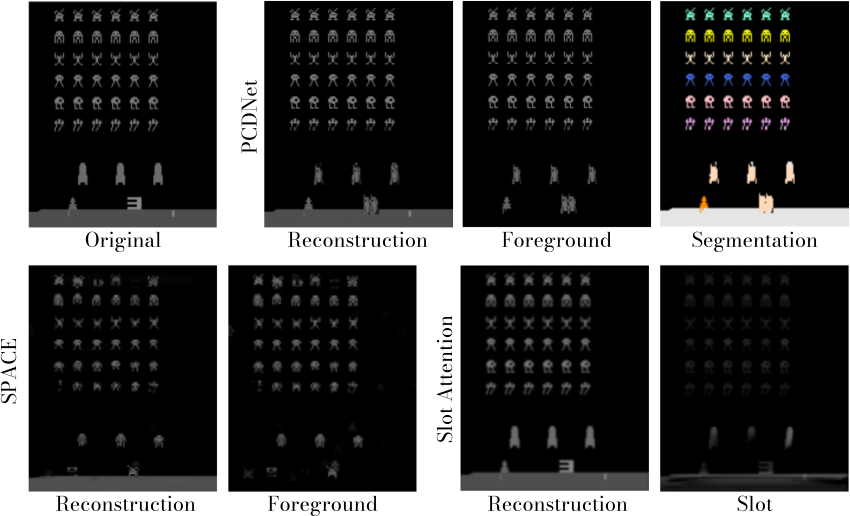}
		\label{fig:ataricompsup1}
	\end{subfigure}
	\hspace{0.009\textwidth}
	\begin{subfigure}{0.495\textwidth}
		\centering
		\includegraphics[width=1.0\linewidth]{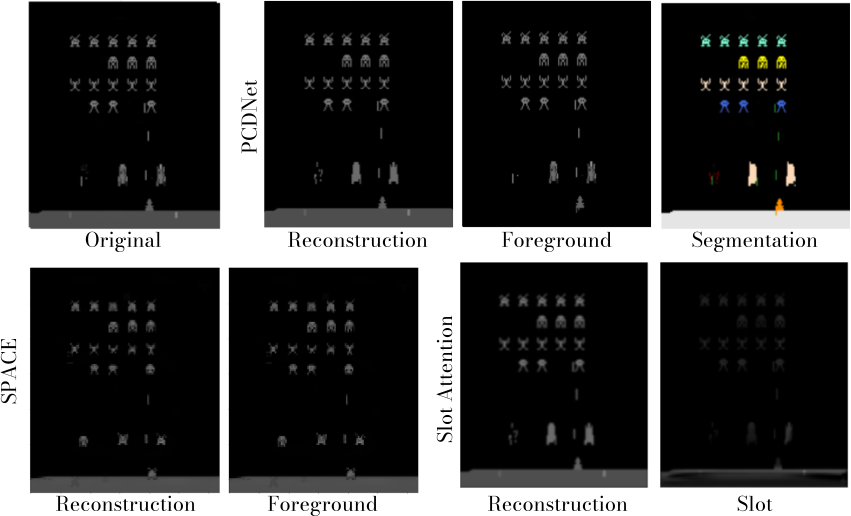}
		\label{fig:ataricompsup2}
	\end{subfigure}
	\vspace{-0.3cm}
	\caption{Additional qualitative comparison on the Space Invaders dataset.}
	\label{fig:ataricompsup}
\end{figure*}

\begin{figure*}[t!]
	\centering
	\includegraphics[width=0.99\linewidth]{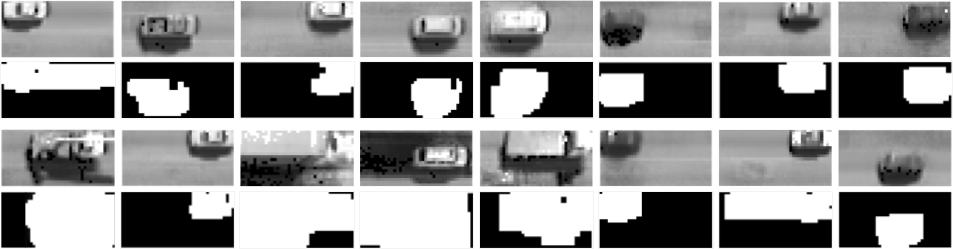}
	\caption{Several vehicle prototypes (top) and their corresponding alpha masks (bottom) learned on the NGSIM dataset.}
	\label{fig:protoscars}
\end{figure*}

\Figure{fig:atarisegment} displays segmentation results obtained by PCDNet on the Space Invaders dataset.
\Figure{fig:ataricompsup} depicts further qualitative comparisons on the Space Invaders dataset between PCDNet, SPACE~\cite{Lin_UnsupervisedObjectOrientedSceneRepresentationViaSpatialAttentionAndDecomposition_2020} and Slot Attention~\cite{Locatello_ObjectCentricLearningWithSlotAttention_2020}.
\Figure{fig:protoscars} depicts several object prototypes and their corresponding alpha
masks learned by PCDNet on the NGSIM dataset.

\vspace*{-0.cm}

\end{document}